\documentclass[conference]{IEEEtran}
\IEEEoverridecommandlockouts \IEEEpubid{\makebox[\columnwidth]{978-1-7281-0858-2/19/\$31.00 $\copyright$ 2019 IEEE \hfill}\hspace{\columnsep}\makebox[\columnwidth]{}}
\usepackage{cite}
\usepackage{amsmath,amssymb,amsfonts}
\usepackage{algorithmic}
\usepackage{graphicx}
\usepackage{textcomp}
\usepackage{xcolor}
\def\BibTeX{{\rm B\kern-.05em{\sc i\kern-.025em b}\kern-.08em
    T\kern-.1667em\lower.7ex\hbox{E}\kern-.125emX}}
\begin{document}

\title{KryptoOracle: A Real-Time Cryptocurrency Price Prediction Platform Using Twitter Sentiments\\
}

\author{\IEEEauthorblockN{Shubhankar Mohapatra}
\IEEEauthorblockA{\textit{Cheriton School of Computer Science} \\
\textit{University of Waterloo}\\
Waterloo, Canada \\
shubhankar.mohapatra@uwaterloo.ca}
\and
\IEEEauthorblockN{Nauman Ahmed}
\IEEEauthorblockA{\textit{Cheriton School of Computer Science} \\
\textit{University of Waterloo}\\
Waterloo, Canada \\
n78ahmed@uwaterloo.ca}
\and
\IEEEauthorblockN{Paulo Alencar}
\IEEEauthorblockA{\textit{Cheriton School of Computer Science} \\
\textit{University of Waterloo}\\
Waterloo, Canada \\
palencar@uwaterloo.ca}

}

\maketitle

\begin{abstract}
Cryptocurrencies, such as Bitcoin, are becoming increasingly popular, having been widely used as an exchange medium in areas such as financial transaction and asset transfer verification. However, there has been a lack of solutions that can support real-time price prediction to cope with high currency volatility, handle massive heterogeneous data volumes, including social media sentiments, while supporting fault tolerance and persistence in real time, and provide real-time adaptation of learning algorithms to cope with new price and sentiment data. In this paper we introduce KryptoOracle, a novel real-time and adaptive cryptocurrency price prediction  platform  based  on  Twitter sentiments. The integrative  and  modular platform is based on (i) a Spark-based architecture which handles the large volume of incoming data in a persistent and fault tolerant way; (ii) an approach that supports sentiment analysis which can respond to  large  amounts  of  natural  language  processing  queries  in real time; and (iii) a predictive method grounded on online learning in which a model adapts  its  weights  to  cope  with  new  prices  and  sentiments. Besides providing an architectural design, the paper also describes the KryptoOracle platform implementation and experimental evaluation. Overall, the proposed platform can  help  accelerate decision-making, uncover new opportunities and provide more timely insights based on the available and ever-larger financial data volume and variety.
\end{abstract}


\section{Introduction}
A cryptocurrency is a digital currency designed to work as a medium of exchange that uses strong cryptography to secure financial transactions, control the creation of additional units, and verify the transfer of assets. They are based on decentralized systems built on block-chain technology, a distributed ledger enforced by a disparate network of computers \cite{investopedia}. The first decentralized cryptocurrency, Bitcoin, was released as open-source software in 2009. After this release, approximately 4000 altcoins (other cryptocurrencies) have been released. As of August 2019, the total market capitalization of cryptocurrencies is \$258 billion, where Bitcoin alone has a market capitalization of \$179 billion \cite{cryptocap}.

Considering the huge market value of these currencies, they have attracted significant attention, where some people consider them as actual currencies and others as investment opportunities. This has resulted in large fluctuations in their prices. For instance in 2017 the value of Bitcoin increased approximately 2000\% from \$863 on January 9, 2017 to a high of \$17,900 on December 15, 2017. However, eight weeks later, on February 5, 2018, the price had been more than halved to a value of just \$6200 \cite{cryptocurrency}.

This high volatility in the value of cryptocurrencies means there is uncertainty for both investors, and for people who intend to use them as an actual currency. Cryptocurrency prices do not behave as traditional currencies and, therefore, it is difficult to determine what leads to this volatility. This in turn makes it a challenge to correctly predict the future prices of any cryptocurrency. To predict these prices, huge heterogeneous data volumes need to be collected from various sources such as blogs, IRC channels and social media. Especially, tweets from highly influential people and mass has significant effects on the price of cryptocurrency \cite{bitcoinTwitter}. However, tweets need to be filtered and their sentiments need to be calculated in a timely fashion to help predict cryptocurrency prices in real time. Furthermore, real-time prediction also calls for real-time updating of learning algorithms, which introduces an additional difficulty. These challenges call for learning platforms based on big data architectures that can not only handle heterogeneous volumes of data but also be fault tolerant and persistent in real time.

In this paper we provide a novel real-time and adaptive cryptocurrency price prediction platform based on Twitter sentiments. The integrative and modular platform copes with the three aforementioned challenges in several ways. Firstly, it provides a Spark-based architecture which handles the large volume of incoming data in a persistent and fault tolerant way. Secondly, the proposed platform offers an approach that supports sentiment analysis based on VADER which can respond to large amounts of natural language processing queries in real time. Thirdly, the platform supports a predictive approach based on online learning in which a machine learning model adapts its weights to cope with new prices and sentiments. Finally, the platform is modular and integrative in the sense that it combines these different solutions to provide novel real-time tool support for bitcoin price prediction that is more scalable, data-rich, and proactive, and can help accelerate decision-making, uncover new opportunities and provide more timely insights based on the available and ever-larger financial data volume and variety. 

The rest of the paper is organized as follows. Section 2 discusses the related work proposed in the literature. Section 3 discusses the design and implementation of KryptoOracle in detail and includes the description of all of its sub-components. Section 4 presents an experimental evaluation, including experimental data, setup and results. Finally, section 5 concludes the paper and describes future work.

\section{Related Work}

In this section we present a brief review of the state of the art related to cryptocurrency price prediction. Related works can be divided into three main categories: (i) social media sentiments and financial markets (including cryptocurrency markets); (ii) machine learning for cryptocurrency price prediction; and (iii) big data platforms for financial market prediction.

The `prospect theory' framed by Daniel Kahneman and Amos Tversky presents that financial decisions are significantly influenced by risk and emotions, and not just the value alone \cite{prospect}. This is further reinforced by other works in \textit{economic psychology} and \textit{decision making} such as \cite{economic} which show that variations in feelings that are widely experienced by people, influence investor decision-making and, consequently, lead to predictable patterns in equity pricing. These insights, therefore, open the possibility to leverage techniques such as sentiment analysis to identify patterns that could affect the price of an entity.

Considering the emergence and ubiquity of media, especially social media, further works have explored how it effects user sentiment and therefore financial markets. Paul Tetlock in \cite{media}, explains how high media pessimism predicts downward pressure on market prices, and unusually high or low pessimism predicts high trading volume. Moreover, Gartner found in a study that majority of consumers use social networks to inform buying decisions \cite{gartner}. This insight has given rise to several research materials which have attempted to find correlations between media sentiments and different financial markets.

The authors in \cite{news} retrieve, extract, and
analyze the effects of news sentiments on the stock market. They develop a sentiment analysis dictionary for the financial sector leading to a dictionary-based sentiment analysis model. With this model trained only on news sentiments, the paper achieved a directional accuracy of 70.59\% in predicting the trends in short-term stock price movement. The authors in \cite{messageboard} use the sentiment of message board comments to predict the stock movement. Unlike other approaches where the overall moods or sentiments are considered, this paper extracts the ‘topic-sentiment’ feature, which represents the sentiments of the specific topics of the company and uses that for stock forecasting. Using this method the accuracy average over 18 stocks in one year transactions, achieved 2.07\% better performance than the model using historical prices only. Similarly, Alan Dennis and Lingyao Yuan collected valence scores on tweets about the companies in the S\&P 500 and found that they correlated with stock prices \cite{emotion}. The authors in \cite{mood}  used a self-organizing fuzzy neural network, with Twitter mood from sentiment as an input, to predict price changes in the DOW Jones Industrial average and achieved a 86.7\% accuracy.

With the recent emergence of cryptocurrencies and the widespread investment in them, has motivated researchers to try to predict their price variations. The authors in \cite{cryptocurrency} predict price fluctuations for three cryptocurrencies: Bitcoin, Litecoin and Ethereum. The news and social media data was labeled based on actual price changes one day in the future for each coin, rather than on positive or negative sentiment. By taking this approach, the model was able to directly predict price fluctuations instead of needing to first predict sentiment. Logistic regression worked best for Bitcoin predictions and the model was able to predict 43.9\% of price increases and 61.9\% of price decreases correctly. A work by Abhraham et al.\ uses Twitter sentiment and google trends data to predict the price of Bitcoin and Ethereum \cite{smucryptocurrency}. The paper uses the tweet volume in addition to the Twitter sentiment to establish a correlation with cryptocurrency price.

KryptoOracle draws greatest inspiration from \cite{kth} and \cite{drabble}. Both works use Twitter sentiments to find correlation with Bitcoin prices. The tweets are cleaned of non-alphanumeric symbols and then processed with VADER (Valence Aware Dictionary and sEntiment Reasoner) to analyze the sentiment of each tweet and classify it as negative, neutral, or positive. The compound sentiment score is then used to establish correlation with the Bitcoin prices over different lag intervals. KryptoOracle builds on what has been discussed above but goes beyond to construct a prediction engine that forecasts Bitcoin prices at specified intervals.

Machine learning has also been employed directly for cryptocurrency price prediction. For instance, the authors in \cite{munim2019next} contribute to the Bitcoin forecasting literature by testing auto-regressive integrated moving average (ARIMA) and neural network auto-regression (NNAR) models to forecast the daily price movement based only on the historical price points. Similarly the author in \cite{spilak2018deep} presents a Neural Network framework to provide a deep machine learning solution to the cryptocurrency price prediction problem. The framework is realized in three instants with a Multi-layer Perceptron (MLP), a simple Recurrent Neural Network (RNN) and a Long Short-Term Memory (LSTM), which can learn long dependencies. In contrast our prediction model in addition to considering the social media influence, also employs online learning to continuously learn from its mistakes and improve itself in the process.

Since our engine is designed to run for an indefinite amount of time and it continuously obtains real-time data, it is inevitable that this will lead to data storage concerns in the long run. Therefore, we treat our objective as a big data problem and employ big data tools to ensure scalability and performance. We take inspiration from \cite{spark} which uses Apache Spark and Hadoop HDFS to forecast stock market trends based on social media sentiment and historical price. Similarly, we leverage the performance of Apache Spark RDDs and the persistence of Apache Hive to build a solution that is fast, accurate and fault-tolerant. To our knowledge KryptoOracle is the first of its kind solution that provides an out of box solution for real-time cryptocurrency price forecasting based on Twitter sentiments while ensuring that the data volume does not become a bottle neck to its performance.

\section{KryptoOracle}
KryptoOracle is an engine that aims at predicting the trends of any cryptocurrency based on the sentiment of the crowd. It does so by learning the correlation between the sentiments of relevant tweets and the real time price of the cryptocurrency. The engine bootstraps itself by first learning from the history given to it and starts predicting based on the previous correlation. KryptoOracle is also capable of reinforcing itself by the mistakes it makes and tries to improve itself at prediction.  In addition, the engine supports trend visualization over time based on records of both incoming data and intermediate results. This engine has been built keeping in mind the increasing data volume, velocity and variety that has been made available and is therefore able to scale and manage high volumes of heterogeneous data.

KryptoOracle has been built in the Apache ecosystem and uses Apache Spark. Data structures in Spark are based on resilient distributed datasets (RDD), a read only multi-set of data which can be distributed over a cluster of machines and is fault tolerant. Spark applications run as separate processes on different clusters and are coordinated by the Spark object also referred to as the SparkContext. This element is the main driver of the program which connects with the cluster manager and helps acquire executors on different nodes to allocate resource across applications. Spark is highly scalable, being 100x faster than Hadoop on large datasets, and provides out of the box libraries for both streaming and machine learning.

\subsection{Architecture}\label{AA}
The growth of the volume of data inspired us to opt for a big data architecture which can not only handle the prediction algorithms but also the streaming and increasing volume of data in a fault tolerant way. 
\begin{figure}[ht]
    \centering
    \includegraphics[width=\columnwidth]{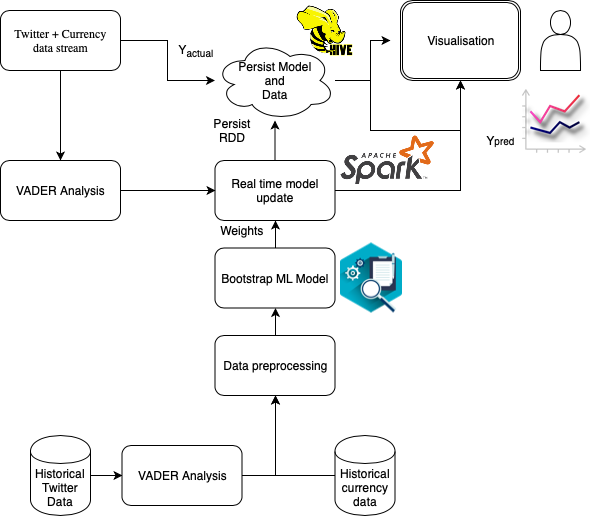}
    \caption{KryptoOracle Architecture}
    \label{fig:architecture}
\end{figure}

Figure \ref{fig:architecture} gives an overview of the architecture design. Central to this design is Apache Spark which acts as an in-memory data store and allows us to perform computations in a scalable manner. This data is the input to our machine learning model for making predictions. To bootstrap our model, we first gather a few days of data and store that in Apache Spark RDDs. Next, we perform computations to construct features from the raw data. All these computations are performed on data that is distributed across multiple Spark clusters and therefore will scale as the data grows continuously.

Once the machine learning model has been bootstrapped, we commence data streaming to get real-time data related to both the social media (in our case, Twitter) and the cryptocurrency. Similar computations are performed on this data to calculate the features and then this new data-point is used to get a future prediction from the model. This computed data-point is then appended to the already existing data in Spark RDDs, obtained from the bootstrap data. Therefore, in addition to making predictions we also keep expanding our data store which allows us to extract holistic visualizations from the data regarding the cryptocurrency market trend and how our own predictions capture that. Moreover, as we discuss later the new data-points are also used to retrain our model.

An important property of this architecture is the persistence of the data and the model. The machine learning model persists itself by storing its weights to disk and loading from it while retraining or reinforcing itself to learn from mistakes. The tweets and cryptocurrency training data is also stored in Apache Hive which provides data warehousing support to read, write and manage distributed datasets directly from disk. This persistence technique helps the whole platform to reset itself without omissions in real time.

Spark RDD has the innate capability to recover itself because it stores all execution steps in a lineage graph. In case of any faults in the system, Spark redoes all the previous executions from the built DAG and recovers itself to the previous steady state from any fault such as memory overload. Spark RDDs lie in the core of KryptoOracle and therefore make it easier for it to recover from faults. Moreover, faults like memory overload or system crashes may require for the whole system to hard reboot. However, due to the duplicate copies of the RDDs in Apache Hive and the stored previous state of the machine learning model, KryptoOracle can easily recover to the previous steady state.

\subsection{Sentiment Analysis}

In KryptoOracle we focus on sentiment analysis on a document level where each tweet is considered as a single document and we intend to determine its sentiment score. In general, there are primarily two main approaches for sentiment analysis: machine learning-based and lexicon-based. Machine learning-based approaches use classification techniques to classify text, while lexicon-based methods use a sentiment dictionary with opinion words and match them with the data to determine polarity. They assign sentiment scores to the opinion words describing how positive or negative the words contained in the dictionary are \cite{vohra2013comparative}. Machine learning-based approaches are inherently supervised and require an adequately large training set for the model to learn the differentiating characteristics of the text corpus. In this paper we choose to forego this training aspect in favour of using a lexicon-based approach. This is because our objective is not to innovate in the natural language processing domain but instead to establish a scalable architecture that is able to capture the relationship between social media sources and financial markets, specifically in the context of the cryptocurrency market. 

To measure the sentiment of each tweet VADER (Valence Aware Dictionary and sEntiment Reasoner) is used \cite{vader}. VADER is a lexicon and rule-based sentiment analysis tool that is specifically attuned to sentiments expressed in social media. When given a text corpus, VADER outputs three valence scores for each sentiment i.e. positive, negative and neutral. A fourth \textit{compound} score is computed by summing the valence scores of each word in the lexicon, adjusted according to the rules, and then normalized to be between -1 (extreme negative) and +1 (extreme positive). To summarize, it is a normalized, weighted composite score. This is the most useful metric for us since it provides a single uni-dimensional measure of sentiment for a given tweet. Therefore, we capture the sentiment of each tweet using the compound score.

However, this score is not the final metric that we use to build our machine learning model. It is quite intuitive that tweets belonging to influential personalities should be assigned more weight since they will have a more significant impact on the price of any cryptocurrency. To capture this relationship the compound score is multiplied by the poster's follower count, the number of likes on the tweet and the retweet count. The final score is calculated with the following equation:
\begin{eqnarray*}
 FinalScore & =
& {} CompoundScore \\ 
& & {} *  UserFollowerCount \\
& & {} * (Likes + 1) * (RetweetCount + 1)
\end{eqnarray*}

The +1 to both the \textit{RetweetCount} and \textit{Likes} ensures that the final score does not become zero if there are no likes or re-tweets for the tweet in subject. \textit{UserFollowerCount} does not have +1 to filter out the numerous bots on Twitter which flood crytocurrency forums. We further normalize the score by taking the root of the final score and multiplying by -1 if the score is negative. This final score belongs to a single tweet and since our prediction scope is for a certain time frame, we sum up all the normalized scores for the different tweets received during that time frame. This summed up score is then used as one of the features for our model to predict the cryptocurrency price for the future time frame.

\subsection{Machine Learning}

An important element of our architecture is the machine learning model, trained to capture the correlation between social media sentiment and a certain metric of the financial market, in our case, the price of cryptocurrency. An essential characteristic of the model is that it should be able to continuously evolve and adjust its weights according to the ever-changing social media sentiments and the volatile cryptocurrency market. We discuss later how we incorporate this in our model design. However, it is worth mentioning that our problem deals with structured data with features related to the social media sentiments and primitive or computed metrics of the cryptocurrency market.

In prediction problems involving unstructured data, ANNs (Artificial Neural Networks) tend to outperform all other algorithms or frameworks. However, when it comes to small-to-medium structured/tabular data like in our case, decision tree based algorithms are currently considered best-in-class. Therefore, we experimented with a few techniques but then ultimately decided to use XGBoost \cite{xgboost} owing to its speed, performance and the quality of being easily re-trainable. XGBoost is under development and will be released to work in PySpark. Therefore, at this moment we choose to deploy the model outside of our Spark framework. For bootstrapping the model, historical data points are exported outside the Spark framework and used to train the model initially. After this, as new real-time data arrives it is processed to create a new data-point of the required features. This data-point is then also exported outside Spark and fed to the machine learning model to obtain a prediction for the future price.

\begin{figure}[!ht]
    \centering
    \includegraphics[width=\columnwidth]{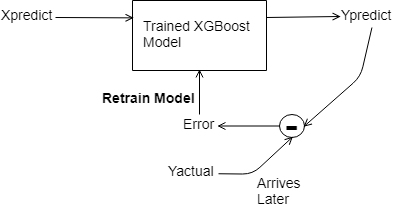}
    \caption{The XGBoost model is retrained on each iteration of the real time stream}
    \label{fig:OnlineLearning}
\end{figure}

To continuously improve the model we employ online learning. The model is saved to disk and after every prediction we wait for the actual price value to arrive. This actual price value is then used to retrain the model as shown in Figure \ref{fig:OnlineLearning}, so that it can learn from the error between the value it had predicted earlier and the actual value that arrived later. In this way the model keeps readjusting its weights to stay up to date with the market trends.

\begin{figure*}[!htb]
  \includegraphics[width=\textwidth, height = 7 cm]{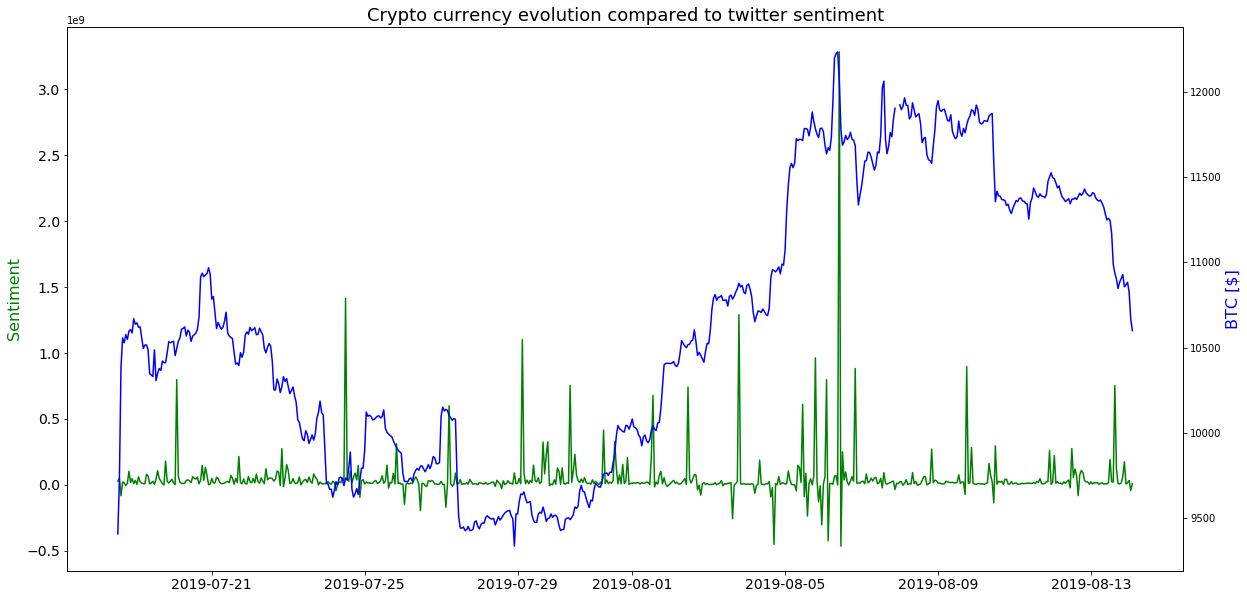}
  \caption{Sentiment scores and Bitcoin prices}
  \label{price}
\end{figure*}
\section{Experimental Evaluation}
We used PySpark v2.3 in Jupyter notebooks with Python 2.7 kernels to code KryptoOracle. The entire source code was tested on a server instance on the SOSCIP cloud with 32 GB RAM, 8 CPUs and 120 GB HDD running on Ubuntu 18.04 over a period of 30 days. The data extraction and correlation codes were taken from ``Correlation of Twitter sentiments with the evolution of cryptocurrencies," which is publicly available\cite{drabble}. The data collected for this experiment was for the Bitcoin cryptocurrency.

\subsection{Data}
The data fed into KryptoOracle is primarily of two types, Twitter data which consists of tweets related to the cryptocurrency and the minutely cryptocurrency value.
\begin{itemize}
    \item Twitter data: We used the Twitter API to scrap tweets with hashtags. For instance, for Bitcoin, the \#BTC and \#Bitcoin tags were used. The Twitter API only allows a maximum of 450 requests per 15 minute and historical data up to 7 days. Throughout our project we collect data for almost 30 days. Bitcoin had about 25000 tweets per day amounting to a total of approximately 10 MB of data daily. For each tweet, the ID, text, username, number of followers, number of retweets, creation date and time was also stored. All non-English tweets were filtered out by the API. We further processed the full tweet text by removing links, images, videos and hashtags to feed in to the algorithm.
    \begin{figure}[ht]
        \centering
        \includegraphics[width=\columnwidth]{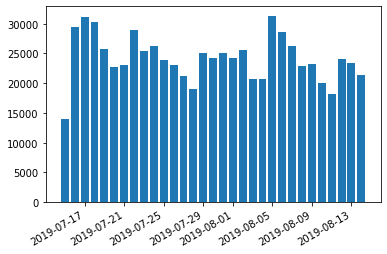}
        \caption{Number of tweets collected per day}
        \label{fig:no_of_tweets}
    \end{figure}
    \item Cryptocurrency data: To obtain cryptocurrency data, the Cryptocompare API \cite{cryptocompare} was used. It provides a free API that provides the 7 day minutely values of any cryptocurrency. The data has several fields: time, open, close, high and low that correspond to the opening, closing, high and low values of the cryptocurrency in that particular time frame in USD.
\end{itemize}

After collecting all the data, we aligned all tweets and cryptocurrecy data by defined time windows of one minute and stored the resulting data into a training data RDD. This training data RDD was further processed as described in the later subsections and then fed into the machine learning algorithm. The same API and structure was also used to stream in real time to KryptoOracle. 

\subsection{Procedure and Results}
We started by collecting Twitter data with hashtags \#Bitcoin and \#BTC for a period of 14 days using Twython, a python library which uses Twitter API to extract tweets using relevant queries. The real time price of Bitcoin was also simultaneously collected using the crytocompare API. The Twitter data was cleaned to remove any hashtags, links, images and videos from the tweets. The sentiment score of each tweet was collected to get the scores as described in the previous section. 
\begin{figure*}[!htb]
  \includegraphics[width=\textwidth ]{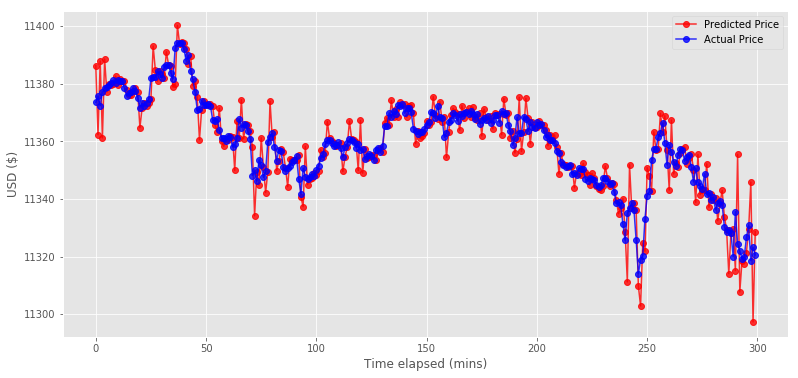}
  \caption{KryptoOracle's predictions}
  \label{prediction}
\end{figure*}
\begin{figure}[htbp]
\begin{minipage}[t]{0.5\linewidth}
    \includegraphics[width=\linewidth]{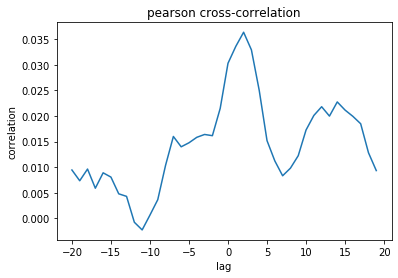}
    \label{pearson}
\end{minipage}%
    \hfill%
\begin{minipage}[t]{0.5\linewidth}
    \includegraphics[width=\linewidth]{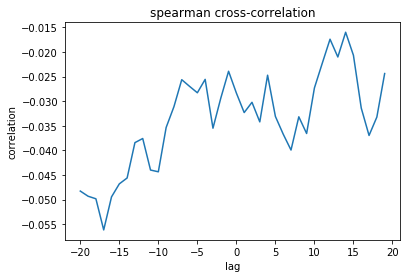}
    \label{spearman}
\end{minipage} 
\caption{Correlation graphs}
\label{correlation}
\end{figure}
To analyze the data, we calculated the Spearman and Pearson correlation between the tweet scores and the Bitcoin prices as shown in Figure \ref{correlation}. The y-axis of the graphs denote the lag in minutes to see if there was any lag between the arrival of tweets and the Bitcoin prices. The trend of the tweet scores and the corresponding Bitcoin prices is captured in Figure \ref{price}. The hourly summed up Twitter sentiments and their corresponding mean bitcoin price for the hour have been plotted in the graph. It can be seen in the figure that some spikes in sentiment scores correspond directly or with some lag with the Bitcoin price. We also noticed that the volume of incoming streaming tweets in the time of a radical change increases, which results in higher cumulative score for the hour.
 
\begin{figure}[!ht]
    \centering
    \includegraphics[width=\columnwidth]{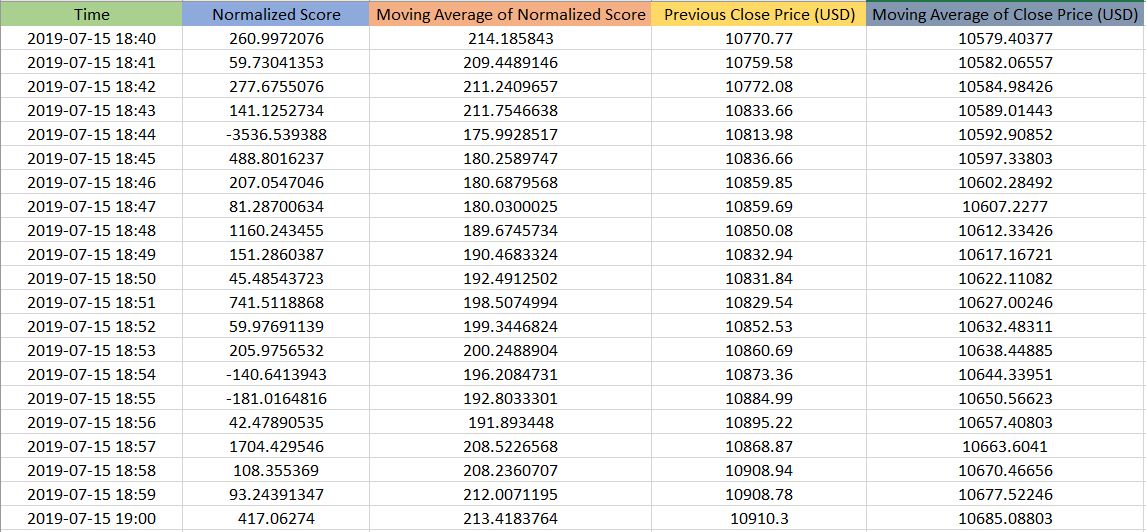}
    \caption{Machine learning Features}
    \label{fig:features}
\end{figure}

The bitcoin price and Twitter sentiment features were not enough to predict the next minute price as they did not capture the ongoing trend. It was therefore important that the historical price of the cryptocurrency was also incorporated in the features so as to get a better prediction for the future. We, therefore, performed some time series manipulation to engineer two new features for our model. The first feature was the \textit{Previous Close Price} that captured the close price of the cryptocurrency in the previous time frame. The next feature was the \textit{Moving Average of Close Price}. This feature was a rolling average of the last 100 time frame close prices and aimed to capture the pattern with which the price was constrained to change. A similar new third feature called \textit{Moving Average of Scores} was designed to capture the rolling average of the last 100 scores. This new feature captured the past sentiment information. With these three additional features combined with the final sentiment score computed in the previous subsection, we got the final training data as shown in Figure \ref{fig:features}.

Once the historical data was stored, all information was fed to the machine learning model. In our experiment, we stored historical data for a month but this can be easily extended as per user requirements. 

 Once the KryptoOracle engine was bootstrapped with historical data, the real time streamer was started. The real-time tweets scores were calculated in the same way as the historical data and summed up for a minute and sent to the machine learning model with the Bitcoin price in the previous minute and the rolling average price. It predicted the next minute's Bitcoin price from the given data. After the actual price arrived, the RMS value was calculated and the machine learning model updated itself to predict with better understanding the next value. All the calculated values were then stored back to the Spark training RDD for storage. The RDD persisted all the data while training and check-pointed itself to the Hive database after certain period of time. 
 
We ran the engine for one day and got an overall root mean square (RMS) error of 10\$ between the actual and the predicted price of Bitcoin. The results for RMS values can be seen below.  
\begin{figure}[!ht]
    \centering
    \includegraphics[width=\columnwidth ]{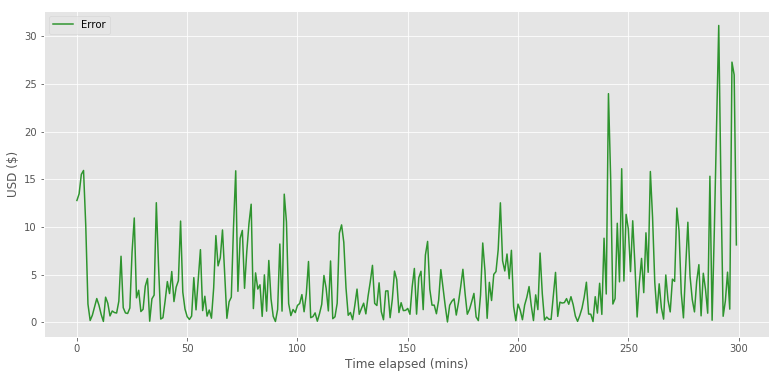}
    \caption{Error in the predicted and actual price measured over 5 hours}
    \label{fig:error}
\end{figure}

Figure \ref{fig:error} shows the RMS error (in USD) for a period of 5 hours at the end of our experiment. The visualization graph at the end of KryptoOracle can be seen in Figure \ref{prediction} which captures the actual price of Bitcoin and the predicted price by KryptoOracle over the same period of 5 hours. The graph shows clearly how KryptoOracle has been able to correctly predict the bitcoin price ahead of 1 minute time. The engine clearly learns from the errors it makes and rewires itself to predict in real-time which can be seen from the adaptive nature of the predicted price graph.

\section{Conclusion and Future Work}
In this paper, we present a novel big data platform that can learn, predict and update itself in real time. We tested the engine on Twitter sentiments and cryptocurrency prices. We envision that this engine can be generalized to work on any real time changing market trend such as stock prices, loyalty towards product/company or even election results. Sentiments in real world can be extracted from not only tweets but also chats from IRC channels, news and other sources such as images and videos from YouTube or TV channels. This implies that the platform can be customized for tasks where the objective is to make predictions based on social media sentiments. In future, we plan to create a front-end for this system which can be used to visually capture the trend and also show historical aggregated data as per user input. Such a front-end could also allow the time window for prediction to be tweaked to predict prices for further ahead in time. 

We understand that crytocurrency prices are influenced by a lot of factors which cannot be captured by Twitter sentiments. Supply and demand of the coin and interest of major investors are two major factors \cite{factors}. To capture these factors one has to add more features to the training data with  inferences from multiple sources such as news, political reforms and macro-financial external factors such as stocks, gold rates and exchange rates. While we performed our experiments, the crytocurrency values did not go through any major changes and thus this engine also needs to be tested with more adverse fluctuations. One way to capture fluctuations can be to trace back to the features that have gone through the major changes and adaptively assign them more weights while training the machine learning model. 

There is also future work related to the machine learning part of the engine. The state of the art time series machine learning algorithms include the modern deep learning algorithms such as RNNs and LSTMs \cite{timeseries}, but unfortunately Spark does not provide deep learning libraries yet. There are some plugins, such as Sparkflow, that facilitate neural network support, but work is also under way to provide Spark with such in-built deep learning support. Currently, Spark also does not have much streaming machine learning support, other than linear regression and linear classification. However, the advent of additional streaming algorithm support in Spark will certainly benefit engines such as KryptoOracle.  








\bibliographystyle{./bibliography/IEEEtran}
\bibliography{./bibliography/test}

\vspace{12pt}

\end{document}